\documentclass[sigconf,nonacm]{acmart}
\AtBeginDocument{%
  }

\setcopyright{acmlicensed}
\copyrightyear{2025}
\acmYear{2025}
\acmConference[C+J '25]{Computation + Journalism Symposium 2025}{December 11--12,
  2025}{Miami, FL}

\begin{document}

\title{Not Wrong, But Untrue: LLM Overconfidence in Document-Based Queries}

\author{Nick Hagar}
\email{nicholas.hagar@northwestern.edu}
\affiliation{%
  \institution{Northwestern University}
  \city{Evanston}
  \state{IL}
  \country{USA}
}

\author{Wilma Agustianto}
\affiliation{%
  \institution{University of Minnesota}
  \city{Minneapolis}
  \state{MN}
  \country{USA}
}

\author{Nicholas Diakopoulos}
\affiliation{%
  \institution{Northwestern University}
  \city{Evanston}
  \state{IL}
  \country{USA}
}

\renewcommand{\shortauthors}{Hagar et al.}

\begin{abstract}
Large language models (LLMs) are increasingly used in newsroom workflows, but their tendency to hallucinate poses risks to core journalistic practices of sourcing, attribution, and accuracy. We evaluate three widely used tools---ChatGPT, Gemini, and NotebookLM---on a reporting-style task grounded in a 300-document corpus related to TikTok litigation and policy in the U.S. We vary prompt specificity and context size and annotate sentence-level outputs using a taxonomy to measure hallucination type and severity. Across our sample, 30\% of model outputs contained at least one hallucination, with rates approximately three times higher for Gemini and ChatGPT (40\%) than for NotebookLM (13\%). Qualitatively, most errors did not involve invented entities or numbers; instead, we observed interpretive overconfidence–models added unsupported characterizations of sources and transformed attributed opinions into general statements. These patterns reveal a fundamental epistemological mismatch: While journalism requires explicit sourcing for every claim, LLMs generate authoritative-sounding text regardless of evidentiary support. We propose journalism-specific extensions to existing hallucination taxonomies and argue that effective newsroom tools need architectures that enforce accurate attribution rather than optimize for fluency.
\end{abstract}




\maketitle

\section{Introduction}
As news organizations increasingly experiment with large language models (LLMs) in reporting and research workflows, journalists and audiences remain concerned about model hallucination---the production of plausible but unsupported statements. Erroneous LLM outputs conflict with journalism’s accuracy norm and accountability practices \cite{cools_uses_2024,porlezza_accuracy_2023,toff_or_2024}, and although model capabilities have advanced, hallucinations persist \cite{rawte_troubling_2023,openai_openai_2025}. This characteristic underscores the need to understand not only \emph{how often} hallucinations occur, but also \emph{what kinds} of errors they might produce in newsroom tasks \cite{diakopoulos_generative_2024,becker_policies_2025}.

A growing technical literature proposes hallucination mitigation strategies to anchor claims in evidence \cite{lewis_retrieval-augmented_2020,fan_survey_2024}, but evaluations of these techniques are typically conducted on general QA, exam-style, or synthetic benchmarks. Evaluation is less common on reporting tasks with heterogeneous document sets and strict provenance expectations \cite{fan_survey_2024}. 

This paper investigates LLM hallucination in the context of document querying for reporting. We assemble a mixed corpus of 300 documents on TikTok litigation and policy in the U.S.---combining news coverage, legal and government materials, and scholarly sources---and evaluate three widely used tools: ChatGPT, Gemini, and NotebookLM. We vary prompt specificity (from broad overviews to precise, document-bound questions) and context size (10, 100, or 300 documents) to mirror realistic choices journalists face when scoping evidence. We annotate outputs at the sentence level using the taxonomy of Rawte et~al.\ to characterize orientations, categories, and degrees of hallucination \cite{rawte_troubling_2023}.

Our findings reveal that hallucinations in journalism tasks reflect an epistemological mismatch between how LLMs and journalists handle evidence. Models exhibited interpretive overconfidence, adding unsupported analysis about document purposes and audiences while transforming attributed claims into universal statements. These patterns appeared across all the tools we evaluated. We propose extensions to existing taxonomies to capture these journalism-specific failure modes and argue that newsroom LLM adoption requires not just better models but different architectures---ones that enforce sourcing requirements rather than optimize for fluency. This work demonstrates why journalistic epistemology \cite{ekstrom_epistemology_2019} poses unique challenges for current LLM design.

\section{Background}
LLMs have seen widespread adoption throughout the news industry, for a range of business and editorial use cases \cite{diakopoulos_generative_2024, becker_policies_2025, cools_uses_2024}. However, the proliferation of generative AI in reporting work remains uneven. Newsrooms often treat LLMs as assistants or tools, rather than autonomous agents, stressing the need for human oversight \cite{becker_policies_2025}. For example, in the context of fact-checking, many high-impact use cases remain speculative \cite{wolfe_impact_2024}. And many news readers respond negatively to the use of AI tools in any capacity, reporting lower trust in stories that leverage LLMs \cite{toff_or_2024}. 

Much of this concern comes from a tension between the journalistic value of \textit{accuracy} and the ongoing prevalence of LLM \textit{hallucinations}. Providing accurate information is a core tenet of journalistic practice, forming the basis of the trust-driven relationship between a news publisher and its audience \cite{noauthor_spjs_nodate}. It is therefore not surprising that, because they explicitly produce inaccurate information, LLM hallucinations are a major source of concern for both journalists and news readers \cite{lipka_americans_2025,cools_uses_2024}.

From a technical perspective, hallucinations emerge from LLMs' ability to generate text that follows common patterns without possessing an understanding of what is true \cite{emslie_llm_2024}. This characteristic results in plausible-sounding responses that do not reflect reality---for example, LLM-fabricated case law that makes its way into arguments \cite{merken_ai_2025}. And while LLM capabilities have increased dramatically over the past five years, hallucinations remain an issue, in some cases even increasing as models become more capable \cite{rawte_troubling_2023,openai_openai_2025}. 

At the same time, researchers have developed a range of strategies to mitigate and better understand LLM hallucinations. Broadly, these approaches fall into three categories:

\textbf{Context.} Models can be grounded with external sources—databases, document collections, or web content—to support their claims with evidence \cite{lewis_retrieval-augmented_2020}. This works well when sources are reliable and complete \cite{fan_survey_2024}. But the approach has limits: Missing information, outdated sources, or poor-quality material can still lead to errors. Models may also make confident claims that go beyond what the sources actually say \cite{fan_survey_2024,lewis_retrieval-augmented_2020}.

\textbf{Prompting and decoding.} Specific instructions can guide models to check their evidence \cite{xiong2024can}, break tasks into steps \cite{trivedi_interleaving_2023,manakul_selfcheckgpt_2023}, or follow strict output formats \cite{geng_grammar-constrained_2023,raspanti_grammar-constrained_2025}. Techniques like having models review their own work or comparing multiple responses can catch some mistakes, though they increase costs and don't always catch subtle errors \cite{gou2024critic,wang2023selfconsistency,geng_grammar-constrained_2023}. Without proper evidence checking, these methods can end up shifting the verification problem onto users, rather than solving it \cite{huang2024large}.

\textbf{Models and tools.} Giving models access to search engines, calculators, or other tools helps them verify information \cite{xiong2024can,nakano_webgpt_2021}. Training on high-quality, sourced data improves accuracy, and citation features help users check claims \cite{fu_mitigating_2025,aly_learning_2024}. But these aren't perfect solutions---they need good sources, clear guidelines, and human review to avoid propagating bad information.

To understand which approaches actually help journalists, we need evaluations that reflect newsroom workflows and standards. This study interrogates hallucination specifically in a reporting context. We evaluate frontier models under prompting strategies and document grounding typical of newsroom workflows, measuring not only error rates but error \emph{types} \cite{rawte_troubling_2023}. We measure not only how often these tools hallucinate on journalism tasks, but what kinds of errors they produce and what that means for newsroom adoption.

\section{Data}
Our analysis focused on the kind of document-based querying that journalists undertake as part of research-based or investigative stories. Rather than attempting to build a comprehensive collection of all relevant materials, we aimed to assemble a corpus that would mirror a representative small-to-medium journalistic document project---substantial enough to reflect real-world research scenarios while remaining manageable in scope. We chose the ongoing legal battle over banning TikTok in the United States as a nuanced, contemporary issue of journalistic interest. To build this corpus, we searched for TikTok ban-related materials across the Washington Post, the New York Times, ProQuest, and Westlaw, retaining the 300 most relevant documents from our search results: 5 academic papers, 150 news articles, and 145 legal documents. These documents are available upon request from the project repository.\footnote{\url{https://osf.io/rbf8t/}}

\section{Methods}
\subsection{Query Approach}
LLM outputs vary greatly depending on the wording of the user’s prompt---particularly its specificity---and the amount and quality of contextual information provided \cite{sclar2024quantifying}. To capture the potential impact of prompt specificity, we designed five queries relevant to the source material ranging from very broad to very specific. At the broadest level, we prompt each LLM to provide dominant arguments for banning TikTok in the United States. At the most specific, we request point-by-point information from particular court cases, with reference to specific testimonies and data. The full text of each query appears in Appendix A. 

To capture the potential impact of context length, we also varied the number of documents that we provided to each LLM, running queries with 10 (3\% of corpus), 100 (33\% of corpus), and 300 (100\% of corpus) documents. We randomly sampled documents from the corpus, with the exception of two that were specifically mentioned in our ``Specific'' query and were included in every sample \cite{noauthor_tiktok_2025,marimow_supreme_2025}. In total, this process produced 15 LLM responses per tool (with the exception of ChatGPT which had only 10, discussed more below) that formed the basis of our analysis. 

\subsection{LLM Tools Used}
We evaluated three tools that represent different approaches to document-based querying:

\textbf{ChatGPT (OpenAI)}---The most widely-used LLM tool, used by 34\% of U.S. adults \cite{mcclain_34_2025}. We used the ``Projects'' feature\footnote{\url{https://help.openai.com/en/articles/10169521-projects-in-chatgpt}} to upload documents. Since projects are limited to 100 documents, we did not evaluate the 300 document condition for this tool. We tested ChatGPT in August 2025, using the version of GPT-5 made available via a ``ChatGPT Plus'' subscription in the UI.

\textbf{Gemini 2.5 Pro (Google)} - A frontier model with one of the largest available context windows (1M tokens), allowing us to include all documents ($\approx$923,000 tokens) directly in prompts. We tested this model in July 2025, using the version of Gemini 2.5 Pro made available via a ``Google AI Pro'' subscription in Gemini.

\textbf{NotebookLM (Google)} - A RAG system that provides citations for its responses. We created dedicated notebooks for each document sample. We tested this tool in July 2025, using the version made available via a ``Google AI Pro'' subscription.

These tools differ in both their underlying models and how they handle documents---ChatGPT and NotebookLM use retrieval, while Gemini processes all documents in-context. This limits direct comparison but reflects the actual tools available to journalists for addressing a practical use-case.

\subsection{Response Annotation and Evaluation}
To ensure that our coding captured the breadth of hallucination phenomena, we adopted the taxonomy proposed by Rawte et al. (2023) \cite{rawte_troubling_2023}. Their framework distinguishes hallucinations along three dimensions: \textit{orientation, category, and degree}. Orientations identify whether a hallucination arises from distortion of factually correct prompts or elaboration on factually incorrect ones. Categories specify the type of error (e.g., misstated numbers, errors in time or duration). Finally, degree marks severity on a three-point scale---mild (questionable superficial details added to otherwise correct information), moderate (clearly false but plausible-seeming facts), and alarming (fundamentally distorted or completely fabricated content). This taxonomy has the advantage of moving beyond binary notions of factuality, providing a finer-grained lens for identifying and comparing hallucination types across systems. All model outputs were annotated by one author. The author reviewed responses sentence by sentence, applying codes according to this taxonomy and flagging additional issues not captured by it.  For hallucinations that did not fit existing taxonomy categories (primarily coded as ``miscellaneous''), we conducted an inductive thematic analysis to identify recurring patterns and develop journalism-specific error typologies.

\section{Results}
\subsection{Hallucination Prevalence}
\begin{figure}
    \centering
    \includegraphics[width=0.9\linewidth]{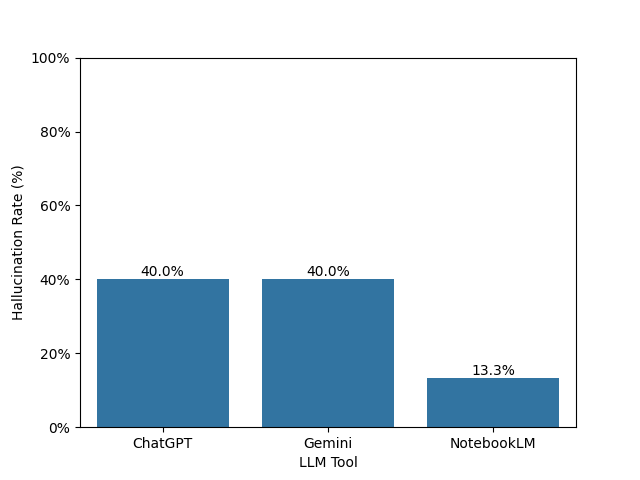}
    \caption{Overall rate of hallucination per tool. Gemini and ChatGPT return the most responses with hallucinations.}
    \label{fig:rates}
\end{figure}

At a high level, 12 out of 40 (30\%) query responses contain any kind of hallucination. However, there is a substantial difference in hallucination rate among tools. As Figure \ref{fig:rates} shows, the rate of hallucinations from Gemini and ChatGPT are approximately three times that of NotebookLM---40\% (4 of 10 for Gemini; 6 of 15 for ChatGPT) versus 13\% (2 of 15 for NotebookLM). This indicates that, while the majority of responses across all tools contain no hallucinations, the choice of tool does make a difference for the same document corpus and query set. 

Hallucinations tend to cluster together. When a response contains any hallucination, it typically contains several: Gemini averages 4 hallucinations per erroneous response, NotebookLM averages 3, and ChatGPT averages 1.5. This clustering suggests systematic issues rather than isolated mistakes.

Of the observed hallucinations, 50\% are classified as moderate. An additional 5 (14\%) are classified as alarming---for example, in one query ChatGPT framed a potential TikTok ban as a reciprocal measure by U.S. lawmakers in response to Chinese policy, a claim entirely absent from the cited source document. This means that, of responses that contain hallucinations, the majority---64\%---introduce factual inaccuracies or tangents. This error rate necessitates careful verification of model outputs against source documents, which may offset the efficiency benefits that motivate LLM adoption.

Looking at the types of hallucinations, almost all observations fall under the ``miscellaneous'' category of this taxonomy. Several hallucinations contained fabricated quotations, and one incorrectly expanded an acronym. Section 5.2 provides a more in-depth exploration of the types of hallucinations observed; however, these results at a high level suggest that existing taxonomies may need expansion to adequately capture the types of hallucinations that appear in reporting workflows. 

Finally, NotebookLM's substantially lower hallucination rate (13\% versus 40\% for both ChatGPT and Gemini) suggests that its RAG implementation with explicit citations provides better grounding than either ChatGPT's Projects feature or Gemini's in-context processing. This advantage is particularly clear in the ``Specific'' query condition, where both ChatGPT and Gemini produced hallucinations regardless of document sample size. The task required extracting information from two particular documents---a condition that NotebookLM's citation-focused RAG system handles more reliably than the other approaches.

\subsection{Qualitative Characteristics of Observed Hallucinations}
Rather than inventing facts or numbers, models exhibited overconfidence through interpretive overreach—making analytical claims the documents didn't support. Models added confident characterizations about document purposes, audiences, and speaker intentions that appeared authoritative but lacked any basis in the actual text. They transformed tentative or attributed statements into definitive claims. This overconfidence took two main forms:

\textbf{Editorializing about source type or audience.} Models frequently added confident metadata about documents that wasn't in the text---claiming an article was ``written for a general audience'' or describing a filing as ``intended for legal practitioners.'' These characterizations sounded plausible but had no basis in the source documents themselves.

\textbf{Attribution drift from opinion to statement.} Models transformed attributed claims into universal facts. A senator's opinion from a hearing became an uncontested statement about TikTok's dangers. A source's concerns quoted in a news article became the model's own assertions. Even when models hedged with phrases like ``is seen as,'' they still failed to specify who was doing the seeing. This tendency not only obscures sources of claims, it also interrupts journalists' ability to assess those sources in-context (e.g., deciding whether a speaker quoted in a document is trustworthy). 

These patterns were not specific to a single tool---both retrieval-augmented and in-context configurations exhibited editorializing and attribution drift. Consistent with Section~\ref{fig:rates}, the higher overall hallucination rate observed for Gemini and ChatGPT meant more opportunities for these behaviors to appear, but the qualitative error modes were shared. For reporting tasks, the problem isn't fabrication but overinterpretation: Models add confident analysis the documents don't support and strip away crucial attribution. The frequent use of the taxonomy's ``miscellaneous'' category for these cases indicates that finer-grained subtypes (e.g., ``unsupported audience/source characterization'' and ``attribution loss'') may be warranted for newsroom evaluations.

\section{Discussion}
Even NotebookLM, the best-performing tool in this study, produced hallucinations in 13\% of responses---a rate that remains concerning for journalists whose credibility depends on accuracy. ChatGPT and Gemini's 40\% hallucination rates make them effectively unusable for unsupervised reporting tasks. But beyond raw error rates, our findings reveal a deeper misalignment between how LLMs and journalists handle uncertainty and evidence.

The hallucinations we observed were characterized by overconfidence. Models didn't invent entities or misstate numbers; they added confident analysis unsupported by sources, characterized documents without evidence, and transformed attributed opinions into declarative statements. Prior work demonstrates LLMs' inability to provide accurate assessments of their confidence \cite{toney-wails_are_2024}; our findings demonstrate a similar shortcoming in the context of journalistic source attribution. This pattern also points to an epistemological incompatibility between algorithmic and journalistic ways of knowing \cite{carlson_automating_2018}. While journalism grounds claims in named sources that undergo editorial scrutiny, LLMs appear to flatten the text in their input documents without assessing (and then conveying) evidentiary support. 

This distinction matters for how journalists work with these tools. The standard advice---verify everything against source documents---remains essential, but our findings nuance what needs verification. Facts and entities are likely accurate; rather, the connective tissue of interpretation, characterization, and attribution requires scrutiny.

Our annotations also suggest ways that hallucination taxonomies should expand to capture journalism's specific failure modes. Existing frameworks focus on fabrication---e.g., invented entities, wrong numbers, or temporal confusion \cite{rawte_troubling_2023}. But the hallucinations we observed in a document-centered journalism task center on interpretation. These aren't explicitly captured by current taxonomies. For newsroom applications, we need evaluation frameworks that treat model overconfidence as seriously as factual errors.

\subsection{Limitations}
Our evaluation focused on a single topic (U.S. TikTok litigation) with a specific document mix. Hallucination patterns may differ for other beats or document types.

The tools themselves imposed constraints that complicate comparison. ChatGPT capped uploads at 100 documents, NotebookLM required manual notebook creation, and Gemini processed everything in-context. In addition, these tools did not surface model parameters (e.g., temperature) that would be tunable via an API call and may affect outputs. These mechanical differences mean we tested not just models but entire systems, making it impossible to isolate whether differences stem from underlying models or document-handling approaches.

Our sample identifies patterns but can't establish precise error rates. The findings should be read as directional, highlighting types of problems journalists will encounter rather than definitive measurements of model performance.

\subsection{Practical Implications}
Our findings suggest specific practices for newsroom LLM use:

\textbf{Tool selection and configuration.} The three-fold difference in error rates indicates tool choice matters more than prompt engineering. NotebookLM's citation requirement provides a structural constraint against interpretive overreach. For sensitive reporting tasks, newsrooms should prioritize tools that enforce source attribution over those optimizing for fluency or speed. When citations aren't built-in, workflows must add them by requiring models to include explicit passage markers for every claim.

\textbf{Verification workflows.} Our results show that fact-checking alone isn't sufficient. Journalists must verify interpretive claims as rigorously as factual ones. This means checking: Does the document actually make this argument or just mention the topic? Who specifically made this claim? Is this characterization of the source (audience, intent, genre) supported by the text itself? These interpretive errors are harder to spot than wrong names or dates because they sound plausible and align with expected narratives.

\textbf{Training and awareness.} Newsrooms need to train staff that LLM errors aren't random---they follow patterns. When models process multiple documents, they excel at surface-level extraction but fail at maintaining attribution chains and distinguishing opinions from facts. Editorial processes should flag any model output that characterizes documents, summarizes positions, or makes comparative claims for heightened scrutiny.

\textbf{System design priorities.} For vendors building journalism tools, our findings suggest that citation infrastructure matters more than model size. Rather than optimizing for benchmark performance, journalism-specific tools should adopt practices that cultivate newsroom trust \cite{hagar_optimizing_2019}---reject outputs that can't be traced to specific sources, maintain speaker attribution through multiple steps, and explicitly mark interpretation versus extraction. Model interfaces might be designed to more easily check the kinds of interpretive errors we observed, accelerating the role of the journalist in checking a response. 

\subsection{Conclusion}
Our evaluation reveals that LLM hallucinations in journalism tasks aren't primarily about fabrication, but overconfidence. Models don't invent facts; they invent interpretations, adding layers of analysis and characterization that sound authoritative but lack textual support. This pattern reflects an epistemological mismatch: While journalism requires claims to trace back to sources, LLMs generate plausible-sounding text regardless of evidentiary basis.

For newsrooms, these findings complicate the standard ``verify everything'' advice. Verification must extend beyond facts to interpretation, checking not just what documents say but whether they actually make the arguments models claim they make. The errors we observed were subtle enough to pass casual review but serious enough to undermine journalistic credibility. As news organizations integrate LLMs into reporting workflows, they need processes that target interpretive overreach and attribution drift.

Ultimately, our results suggest that the core challenge is making LLMs more epistemologically aligned with journalism. Until models can maintain clear provenance chains and distinguish reporting from analysis, they remain tools that require intensive supervision rather than trusted partners in the newsroom.

\begin{acks}
This work was conducted with support from the Knight Foundation.
\end{acks}

\bibliographystyle{ACM-Reference-Format}
\bibliography{sample-base}

\appendix

\section{LLM Queries}
\begin{itemize}
    \item \textbf{Very Broad:} What are the most dominant arguments for banning TikTok in the United States?
    \item \textbf{Broad:} Summarize the discussion in regards to TikTok in the United States from the top 5 comprehensive documents?
    \item \textbf{Medium:} What were the Executive/national order argued in terms of TikTok and national threat in the court cases?
    \item \textbf{Specific:} Combining 2 documents, 24-656 Tiktok Inc. v. Garland (01/17/2025) and The Washington Post article The supreme court takes up the TikTok ban. Explain the similarities or differences of the case?
    \item \textbf{Very specific:} What are the most dominant arguments for banning TikTok in the United States point by point, present in the 24-656 Tiktok Inc. v. Garland (01/17/2025) court cases, with reference to specific testimonies and data? (give me the page number for these testimonies). 
\end{itemize}

\end{document}